\documentclass[10pt,twocolumn,letterpaper]{article}

\usepackage{cvpr}              

\usepackage{graphicx}
\usepackage{amsmath}
\usepackage{amssymb}
\usepackage{booktabs}
\usepackage{makecell}
\usepackage{multirow}
\usepackage{siunitx}
\usepackage[version=4]{mhchem}
\usepackage{bbding}
\usepackage{diagbox}
\usepackage{xcolor}
\usepackage{microtype}
\usepackage{colortbl}
\usepackage{tabulary}

\usepackage[pagebackref,breaklinks,colorlinks]{hyperref}

\usepackage[capitalize]{cleveref}
\crefname{section}{Sec.}{Secs.}
\Crefname{section}{Section}{Sections}
\Crefname{table}{Table}{Tables}
\crefname{table}{Tab.}{Tabs.}

\usepackage[pagebackref,breaklinks,colorlinks]{hyperref}
\newcommand\blfootnote[1]{%
  \begingroup
  \renewcommand\thefootnote{}\footnote{#1}%
  \addtocounter{footnote}{-1}%
  \endgroup
}

\newcommand{\urlNewWindow}[1]{\href[pdfnewwindow=true]{#1}{\nolinkurl{#1}}}

\definecolor{citecolor}{HTML}{0071bc}
\hypersetup{
  breaklinks   = true,
  colorlinks   = true, %
  citecolor    = citecolor %
}

\def\cvprPaperID{1922} 
\def\confName{CVPR}
\def\confYear{2023}

\begin{document}

\newcommand\embed[0]{\boldsymbol{\ell}_i}
\newcommand\nerf[0]{\mathbf{F}}
\newcommand\stepfunc[0]{h}
\newcommand\mulstepfunc[0]{\mathbf{H}}

\newcommand{\todo}{{\color{red}{[TODO] }}}
\newcommand{\sida}[1]{\textcolor{cyan}{[Sida: #1]}}
\newcommand{\tocite}[1]{\textcolor{red}{[TOCITE]}}
\newcommand{\ruojin}[1]{\textcolor{blue}{[Ruojin: #1]}}
\newcommand{\haotong}[1]{\textcolor{cyan}{[Haotong: #1]}}
\newcommand{\noah}[1]{\textcolor{orange}{[Noah: #1]}}

\newenvironment{packed_enum}{
\begin{enumerate}
  \setlength{\itemsep}{1pt}
  \setlength{\parskip}{2pt}
  \setlength{\parsep}{0pt}
}{\end{enumerate}}

\newenvironment{packed_item}{
\begin{itemize}
  \setlength{\itemsep}{1pt}
  \setlength{\parskip}{2pt}
  \setlength{\parsep}{0pt}
}{\end{itemize}}

\newcommand\flickrcc[1]{\small{Photos by Flickr users #1 under \href{https://creativecommons.org/licenses/by/2.0/}{CC BY}}}

\newcommand{\beginsupplement}{%
        \setcounter{table}{0}
        \renewcommand{\thetable}{S\arabic{table}}%
        \setcounter{figure}{0}
        \renewcommand{\thefigure}{S\arabic{figure}}%
     }

\definecolor{colorfirst}{rgb}{.866,.945, 0.831} 
\definecolor{colorsecond}{rgb}{1, 0.98, 0.83} 
\definecolor{colorthird}{rgb}{0.76, 0.87, 0.92} 

\newcommand{\cellfirst}{\cellcolor{colorfirst}}
\newcommand{\cellsecond}{\cellcolor{colorsecond}}
\newcommand{\cellthird}{\cellcolor{colorthird}}

\newcommand{\textfirst}{\colorbox{colorfirst}}
\newcommand{\secondtext}{\colorbox{colorsecond}}
\newcommand{\thirdtext}{\colorbox{colorthird}}

\title{Neural Scene Chronology}


\author{
{Haotong Lin}$^{1,2*}$
\quad
Qianqian Wang$^{2}$
\quad
Ruojin Cai$^{2}$
\quad
Sida Peng$^1$ \\[1.mm]
\quad
Hadar Averbuch-Elor$^3$
\quad
Xiaowei Zhou$^{1\dagger}$
\quad
Noah Snavely$^2$
\\[1.mm]
$^1$Zhejiang University
\quad
$^2$Cornell University
\quad
$^3$Tel Aviv University
}


\twocolumn[\maketitle\vspace{-3em}
\begin{center}
\includegraphics[width=\linewidth]{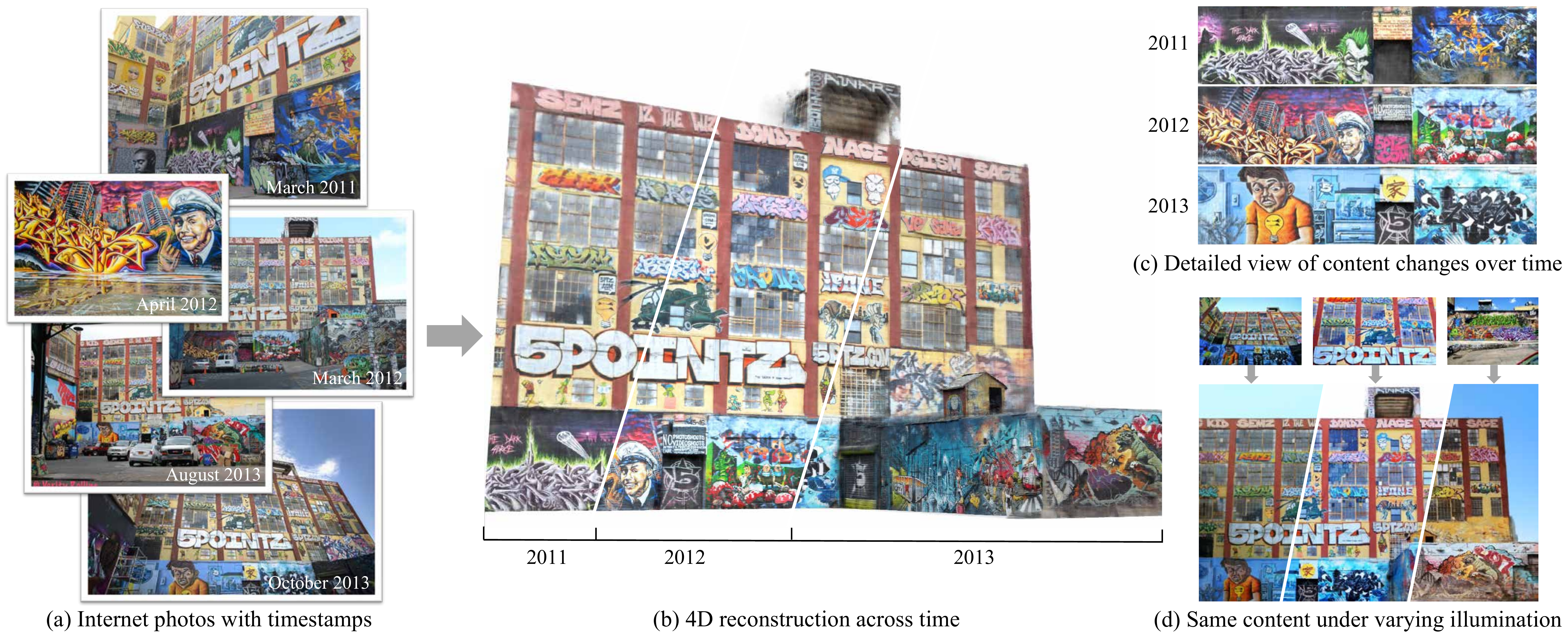}
\end{center} \vspace{-1.5em}
\captionof{figure}{\textbf{Chronology reconstruction.} 
Given timestamped Internet photos \textbf{(a)} of a landmark that has changed significantly over the years (e.g., \href{https://en.wikipedia.org/wiki/5_Pointz}{\textit{5Pointz, NYC}}, the collective graffiti art project shown above), our method can reconstruct a time-varying 3D model, and render photo-realistic images \textbf{(b)} with independent control of viewpoint, time \textbf{(c)} and illumination \textbf{(d)}. \flickrcc{Ryan Brown, DaniGMX, DJ Leekee, Diff Graff, Lee Smith, James Prochnik, Verity Rollins Photo}.
}
\label{fig:teaser}
\bigbreak]

\blfootnote{The authors from Zhejiang University are affiliated with the State Key Lab of CAD\&CG. $^*$This work was done when Haotong Lin was in a remote internship at Cornell University. $^\dagger$Corresponding author: Xiaowei Zhou.}

\begin{abstract}
In this work, we aim to reconstruct a time-varying 3D model, capable of rendering photo-realistic renderings with independent control of viewpoint, illumination, and time, from Internet photos of large-scale landmarks.
The core challenges are twofold. First, different types of temporal changes, such as illumination and changes to the underlying scene itself 
(such as replacing one graffiti artwork with another) are entangled together in the imagery.
Second, scene-level temporal changes are often discrete and sporadic over time, rather than continuous.
To tackle these problems, we propose a new scene representation equipped with a novel temporal step function encoding method that 
can model discrete scene-level content changes as piece-wise constant functions over time. Specifically, we represent the scene as a space-time radiance field with a per-image illumination embedding, where temporally-varying scene changes are encoded using a set of learned step functions. To facilitate our task of chronology reconstruction from Internet imagery, we also collect a new dataset of four scenes that exhibit various changes over time. We demonstrate that our method exhibits state-of-the-art view synthesis results on this dataset, while achieving independent control of viewpoint, time, and illumination. 
Code and data are available at \url{https://zju3dv.github.io/neusc/}.
\end{abstract}

\section{Introduction}

If we revisit a space we once knew during our childhood, it might not be as we remembered it. The buildings may have weathered, or have been newly painted, or may have been replaced entirely.
Accordingly, there is no such thing as a single, authoritative 3D model of a scene---only a model of how it existed at a given instant in time.
For a famous landmark, Internet photos can serve as a kind of chronicle of that landmark's state over time, if we could organize the information in those photos in a coherent way. For instance, if we could reconstruct a time-varying 3D model, then we could revisit the scene at any desired point in time.

In this work, we explore this problem of \emph{chronology reconstruction}, revisiting the work on Scene Chronology from nearly a decade ago~\cite{matzen2014scene}.
As in that work, we seek to use Internet photos to build a 4D model of a scene, from which we can dial in any desired time (within the time interval where we have photos). However, the original Scene Chronology work was confined to reconstructing planar, rectangular scene elements, leading to limited photo-realism.
We can now revisit this problem with powerful neural scene representations, inspired by methods such as NeRF in the Wild~\cite{martinbrualla2020nerfw}. 
However, recent neural reconstruction methods designed for Internet photos assume that the underlying scene is static, which works well for landmarks with a high degree of permanence, but fails for other scenes, like New York's Times Square, that feature more ephemeral elements like billboards and advertisements.

However, we find that adapting neural reconstruction methods~\cite{martinbrualla2020nerfw} to the chronology reconstruction problem has many challenges, and that straightforward extensions do not work well. 
For instance, augmenting a neural radiance field (NeRF) model with an additional time input $t$, and fitting the resulting 4D radiance field to a set of images with timestamps yields temporally oversmoothed models, where different scene appearances over time are blended together, forming ghosted content; such a model \emph{underfits} the temporal signal.
On the other hand, applying standard positional encoding~\cite{mildenhall2020nerf} to the time input \emph{overfits} the temporal signal, conflating transient appearance changes due to factors like illumination with longer-term, sporadic changes to the underlying scene itself.

Instead, we seek a model that can disentangle transient, \emph{per-image} changes from longer-term, \emph{scene-level} changes, and that allows for independent control of viewpoint, time, and illumination at render-time. Based on the observation that scene-level content changes are often sudden, abrupt ``step function''-like changes (e.g., a billboard changing from one advertisement to another), we introduce a novel encoding method for time inputs that can effectively model piece-wise constant scene content over time, and pair this method with a per-image illumination code that models transient appearance changes.
Accordingly, we represent 4D scene content as a multi-layer perceptron (MLP) that stores density and radiance at each space-time $(x, y, z, t)$ scene point, and takes an illumination code as a side input. 
The time input $t$ to this MLP is encoded with our proposed \emph{step function encoding} that models piecewise constant temporal changes. When fit to a set of input images, we find that our representation can effectively factor different kinds of temporal effects, and can produce high-quality renderings of scenes over time. 

To evaluate our method, we collect a dataset of images from Flickr and calibrate them using COLMAP, resulting in 52K successfully registered images.
These photos are sourced from four different scenes, including dense tourist areas, graffiti meccas, and museums, building upon the datasets used in Scene Chronology. 
These scenes feature a variety of elements that change over time, including billboards, graffiti art, and banners. 
Experiments on these scenes show that our method outperforms current state-of-the-art methods and their extensions to space-time view synthesis~\cite{martinbrualla2020nerfw,chen2022hallucinated}.
We also present a detailed ablation and analysis of our proposed time encoding method.

In summary, our work makes the following contributions:
\begin{packed_item}
    \item To the best of our knowledge, ours is the first work to achieve photo-realistic chronology reconstruction, allowing for high-quality renderings of scenes with controllable viewpoint, time, and illumination.
    \item We propose a novel encoding method that can model
    abrupt content changes without overfitting to transient factors.
    This leads to a fitting procedure that can effectively disentangle illumination effects from content changes in the underlying scene.
    \item We benchmark the task of chronology reconstruction from Internet photos and make our dataset and code available to the research community.
\end{packed_item}

\section{Related Work}

\noindent \textbf{3D/4D reconstruction from Internet photos.}
The typical 3D reconstruction pipeline for Internet photos involves first recovering camera poses and a sparse point cloud using Structure from Motion  (SfM) methods~\cite{snavely2006photo,agarwal2011building,schonberger2016structure,schaffalitzky2002multi,snavely2008modeling}, then computing a dense reconstruction using Multi-View Stereo (MVS) algorithms~\cite{goesele2007multi,furukawa2010towards,frahm2010building,schonberger2016pixelwise}.
However, these methods assume the scene to be largely static, and are unable to produce coherent models for scenes with large-scale appearance changes over time.
To extend these methods to achieve 4D reconstruction, Schindler and Dellaert developed a method that takes photos of a city over time, and reasons probabilistically about visibility and existence of objects like buildings that may come and go across decades~\cite{schindler2010probabilistic}.
Most related to our work, Scene Chronology extends MVS methods~\cite{shen2013accurate} to 4D by clustering reconstructed 3D points into space-time cuboids~\cite{matzen2014scene}.
However, it can only reconstruct and render planar regions, leading to limited photo-realism.
To handle more complex geometry, Martin-Brualla \etal represent scene geometry using time-varying depth maps, allowing their method to generate high-quality time-lapse videos~\cite{martin2015time,martin20153d}. 
However, this depth map--based representation limits the range of camera viewpoints their method can synthesize. In our work, we tackle these challenges and devise a new method that can handle large-scale scenes with complex geometry, and can generate large camera motions. 

\medskip
\noindent \textbf{Novel view synthesis.}
Early methods achieve novel view synthesis through light field interpolation~\cite{gortler1996lumigraph, levoy1996light, davis2012unstructured} or image-based rendering~\cite{zitnick2004high, chaurasia2013depth,flynn2016deepstereo, kalantari2016learning}.
Recently, neural scene representations~\cite{niemeyer2020differentiable,park2019deepsdf,mescheder2019occupancy,sitzmann2019deepvoxels,mildenhall2019local,tucker2020single,wizadwongsa2021nex} have shown unprecedented view synthesis quality. Of particular interest is NeRF~\cite{mildenhall2020nerf}, which represents radiance fields using a multi-layer perceptron (MLP) and achieves impressive rendering results.
Many works~\cite{li2020neural,xian2021space,pumarola2020d,park2021nerfies,park2021hypernerf,lin2022efficient,zhang2021editable,li2022neural,wang2021neural,fang2022fast,peng2023representing,instant_nvr,kplanes_2023,li2023dynibar} extend NeRF to model dynamic scenes with moving objects given a monocular or multi-view video as input. In our work, we focus on a different type of 4D view synthesis problem that involves modeling unstructured Internet photo collections capturing scenes that exhibit substantial appearance changes over time.

\medskip
\noindent \textbf{Neural rendering from Internet photos.}
One challenge of rendering from Internet photos is handling varying, unknown illumination present in the image collection. 
Recently, several neural rendering methods demonstrate promising results on rendering static landmarks while allowing for control of illumination effects~\cite{meshry2019neural,li2020crowdsampling}.
In particular, NeRF-W~\cite{martinbrualla2020nerfw} conditions a reconstructed neural radiance field on a learnable per-image illumination vector, thereby factoring out per-image illumination effects.
Chen \etal~\cite{chen2022hallucinated} propose a CNN module for predicting an illumination vector from an image, enabling transfer of illumination from unseen images to the model. Sun \etal~\cite{sun2022neuconw} build on NeRF-W to reconstruct 3D meshes from a collection of Internet photos.
Finally, Zhao and Yang \etal~\cite{neural_outdoor_rerender} and Rudnev \etal~\cite{rudnev2022nerfosr} enable outdoor scene relighting based on neural radiance fields. 
However, these methods are limited to primarily static landmarks like the Brandenburg Gate, and cannot handle scenes with substantial changes over time like Times Square.

\medskip
\noindent \textbf{Modeling temporal signals.} 
One useful type of data for modeling temporal signals is time-lapse videos from stationary cameras, which provide 
organized visual information for scene understanding and factorization. Many previous methods~\cite{liu2020learning,li2018learning,sunkavalli2007factored} show how to factor temporally-varying factors (e.g., illumination) from permanent scene factors (e.g., geometry and reflectance) from time-lapse videos. 
More recently, \cite{harkonen2022tlgan} introduces a method that disentangles time-lapse sequences in a way that allows separate, after-the-fact control of overall trends, cyclic effects, and random effects in the images. 
In our work, we focus on a more challenging setup, where our input is unstructured Internet photos from different viewpoints, and where we aim to synthesize novel views in addition to factorizing different temporal components.

\section{Method}
\begin{figure}
    \centering
    \includegraphics[width=\columnwidth]{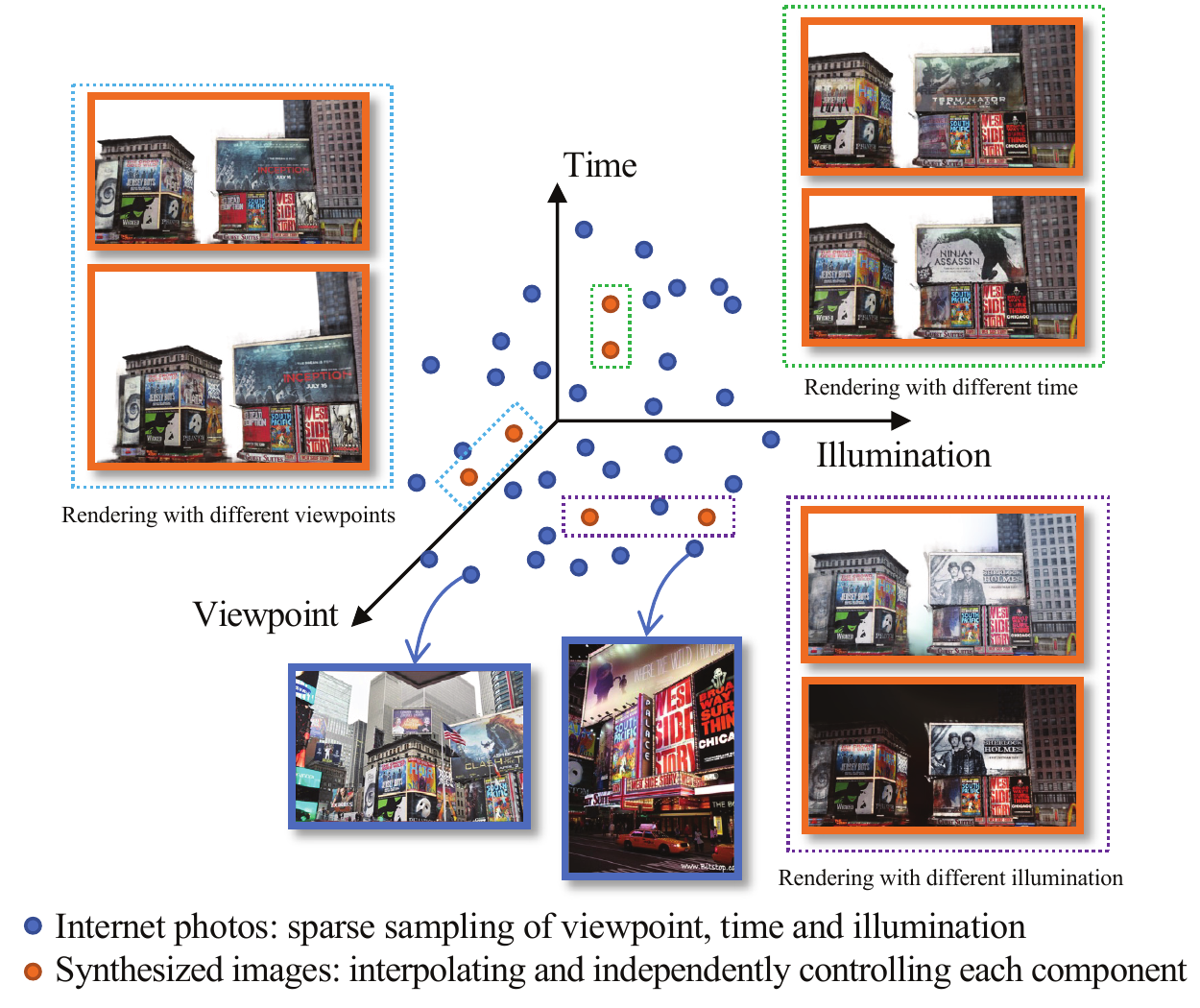}
    \vspace{-5mm}
    \caption{\textbf{Problem illustration.} 
    Given an Internet photo collection of a scene, each image can be thought of as a sample in a high-dimensional space consisting of entangled information, including viewpoint, time, and lighting effects.
    The photo collection represents a sparse sampling of this space. 
    Our goal is to recover a 4D scene representation from this sparse sampling, and to enable interpolation in this high-dimensional space with disentanglement and independent control of viewpoint, time, and lighting.
    }
    \vspace{-3mm}
    \label{fig:task_overview}
\end{figure}

The input to our method is a collection of Internet photos of a landmark (e.g., Times Square) with known timestamps and camera poses. Our goal is to recover a 4D scene representation that can be used to render photo-realistic images of that scene with independently controlled viewpoint, time, and lighting effects as illustrated in \cref{fig:task_overview}. This is a challenging problem because different kinds of temporal changes, including scene content changes (changes to the scene itself) and lighting variation (e.g., time of day) are entangled in each image, but must be disentangled in the scene representation to enable independent control over each temporal component. Furthermore, content changes in our target scenes often happen suddenly, meaning that the scene representation must be able to model discrete, sporadic changes over time.

To tackle this problem, we propose a new 4D scene representation that can disentangle viewpoint, lighting effects, and time. 
Our key observation is that the scene content often changes less frequently over time and remains nearly constant in-between changes, whereas illumination changes much more frequently and sometimes dramatically. 
For example, the graffiti in \textit{5Pointz} (see Fig.~\ref{fig:teaser}) may only be replaced every few months, 
but illumination can change over the course of a few hours.
Motivated by this observation, we model illumination variation with a per-image illumination embedding, and model the underlying 4D scene content using an MLP with time as input. 
We introduce our scene representation in \cref{sec:scene_representation}.
To model piece-wise constant temporal content with abrupt transitions, we propose a novel encoding method in \cref{sec:step_func} that utilizes the behavior of the step function, i.e., remaining consistent given a continuous input, while allowing abrupt changes at transition points.
\begin{figure*}
    \centering
    \includegraphics[width=\linewidth]{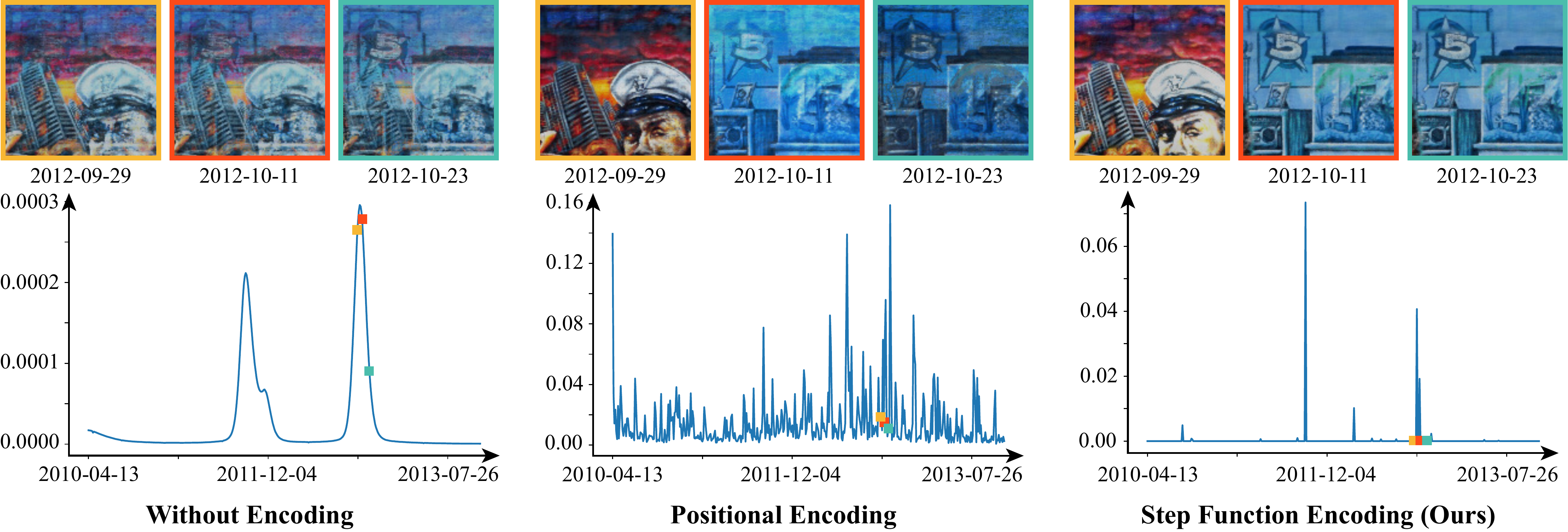}
    \caption{\textbf{Visual ablation of the proposed time encoding.} Given a video rendered at a fixed viewpoint through several years, we compute the Mean Squared Error (MSE) between every pair of consecutive frames. The vertical and horizontal axes represent MSE and time, respectively. For a scene with sporadic changes, we expect to see infrequent ``deltas'' in this MSE plot. We observe that our method indeed 1) recovers the transition points of scene changes and 2) stays consistent (zero MSE) at other times. With no time encoding, 1) is violated, leading to blended, ghosted visual content (top left), and with  positional encoding, 2) is violated, leading to temporal flicker (top middle). Please see our supplemental video for animated results.}
    \label{fig:main_ablation}
    \vspace{-3mm}
\end{figure*}

\subsection{4D Reconstruction from Internet Photos}
\label{sec:scene_representation}
Given a posed image collection $\{\mathcal{I}_i\}_{i=1}^{N}$ with timestamps $\{t_i\}_{i=1}^{N}$, we represent the 4D scene as a time-varying neural field.
To disentangle changes to the underlying scene from the varying and unknown per-image illumination, 
each image $\mathcal{I}_i$ is assigned a learnable illumination embedding vector $\embed$, which is meant to encode the illumination present in that view.
Formally, given a space-time point $(\mathbf{x}, t_i)$ with 3D spatial coordinate $\mathbf{x}$ and time $t_i$, along with an 
illumination code $\embed$ and view direction $\mathbf{d}$, we use an MLP denoted by $\nerf$ to encode 
its radiance $\mathbf{c}$ 
and volume density $\sigma$ as follows:
\begin{equation}
    \mathbf{c}, \sigma = \nerf(\mathbf{x}, t_i, \embed, \mathbf{d}).
\label{equ:simp_model}
\end{equation}

Following NeRF~\cite{mildenhall2020nerf}, the input spatial coordinates $\mathbf{x}$ and ray direction $\mathbf{d}$ are mapped to higher-dimensional vectors via a fixed positional encoding function. 
For simplicity, we assume that the scene geometry is mostly constant, and only the appearance changes over time, but our method could also be extended to handle time-varying geometry.
Therefore the model in \cref{equ:simp_model} can be divided into a static, time-invariant geometric model and a time-aware appearance model, 
denoted by $\nerf_\text{geo}$ and $\nerf_\text{app}$, respectively:
\begin{equation}
    \mathbf{v},\sigma  = \nerf_\text{geo}(\mathbf{x}),
\end{equation}
\begin{equation}
    \mathbf{c}= \nerf_\text{app}(\mathbf{x}, \mathbf{v}, t_i, \embed, \mathbf{d}).
\end{equation}
$\nerf_\text{geo}$ models static geometry, and is parameterized by just the input 3D position $\mathbf{x}$, while $\nerf_\text{app}$ models time-dependent appearance, and depends on the space-time point $(\mathbf{x}, t_i)$, illumination embedding $\embed$, view direction $\mathbf{d}$, and the intermediate geometry feature vector $\mathbf{v}$ produced by $\nerf_\text{geo}$.
Please refer to the supplementary material for additional details about the model architecture.

From this scene representation, we can render images using volume rendering, and optimize the scene representation by comparing these rendered images to the known input views via an image reconstruction loss.
Specifically, given an input image $\mathcal{I}_i$ with timestamp $t_i$, we compute the color of a ray $\mathbf{r}(s)=\mathbf{o}+s\mathbf{d}$, emitted from the camera center $\mathbf{o}$ through a given pixel in direction $\mathbf{d}$ as follows:
We use stratified sampling to sample a set of quadrature points $\{s_k\}_{k=1}^K$ between $s_n$ and $s_f$, the near and far bounds along the ray. 
Then, given the illumination embedding $\embed$ and timestamp $t_i$ of image $\mathcal{I}_i$, we can compute the color $\mathbf{c}(s_k,t_i,\embed)$ and density $\sigma(s_k)$ of each sample $s_k$ given our scene representation.
We then accumulate these points using volume rendering, as in NeRF~\cite{mildenhall2020nerf}, 
yielding the expected color $\hat{\mathbf{C}}(\mathbf{r},t_i,\embed)$:
\begin{equation}
\hat{\mathbf{C}}(\mathbf{r},t_i,\embed)=\sum_{k=1}^{K} T(s_k) \alpha(\sigma(s_k)\delta_k) \mathbf{c}(s_k,t_i,\embed),
\end{equation}
\begin{equation}
\text{where} {\quad} T(s_k)=\text{exp}\left(-\sum_{k'=1}^{k-1}\sigma(s_{k'})\delta_{k'}\right),
\end{equation}
where $\alpha(x)=1-\text{exp}(-x)$, and $\delta_k=s_{k+1}-s_k$.
We minimize the sum of squared error between the rendered and ground truth pixels:
\begin{equation}
\mathcal{L} =\sum_{(\mathbf{r},i)\in\Omega}\|\mathbf{C}_i(\mathbf{r})-\hat{\mathbf{C}}(\mathbf{r},t_i,\embed)\|_2^2 ,
\end{equation}
where $\mathbf{C}_i(\mathbf{r})$ is the observed pixel color in image $\mathcal{I}_i$ with timestamp $t_i$, and $\Omega$ is the set of all the sampled pixels from the image collection $\{\mathcal{I}_i\}_{i=1}^{N}$.

\subsection{Step Function Encoding for Time Input}
\label{sec:step_func}
The method described above serves as a baseline to model a 4D scene from Internet photos.
However, we found that this baseline cannot model temporal changes in the target scene well.
Specifically, temporal appearance changes in man-made scenes are often abrupt, such as a new billboard or sign in Times Square, or a new graffiti artwork in an art mecca like 5Pointz. 
In contrast, the baseline above tends to average over temporal content changes, resulting in a cross-fade transition in time between two appearance states, rather than a sharp, sudden transition. \cref{fig:main_ablation} shows an example where this baseline (denoted ``without encoding'' in the figure) produces a ghosted blend of two temporally consecutive graffiti artworks. 
This finding is consistent with NeRF's observation that standard coordinate inputs cannot model high-frequency signals~\cite{mildenhall2020nerf}.
To address this issue, NeRF uses positional encoding to map input spatial coordinates to a high-frequency signal.
However, we found that applying positional encoding to the time input causes the network to not only fit the underlying appearance changes in the scene, but also overfit to per-image lighting effects. 
In other words, it fails to disentangle these two components and leads to severe flickering artifacts over time, as shown in \cref{fig:main_ablation}.

\begin{figure}
    \centering
    \includegraphics[width=\columnwidth]{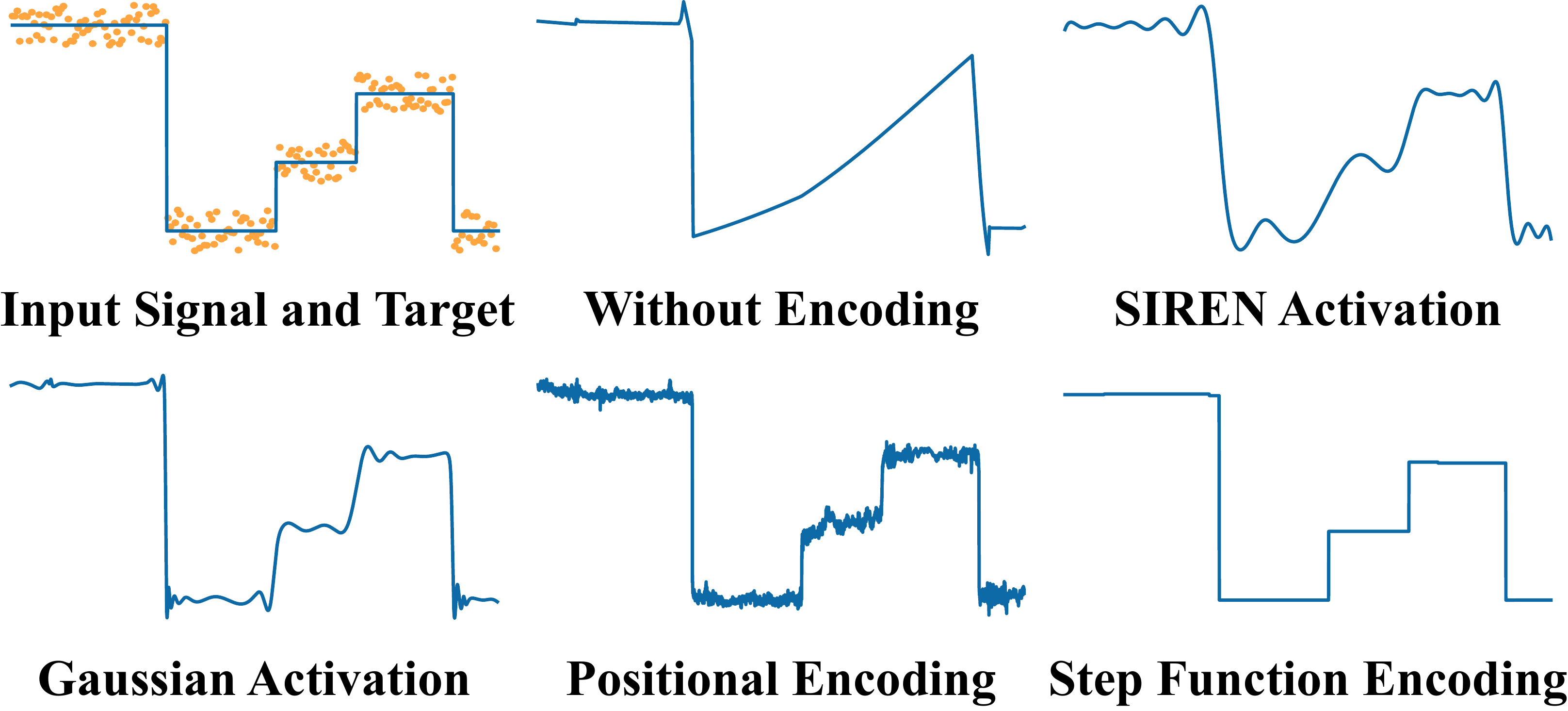}
    \caption{\textbf{Fitting an MLP to a noisy piecewise constant 1D signal.} Given the noisy orange data points (uniform noise), we aim to recover the clean blue curve shown in the upper left plot. We present the results of fitting the input data using an MLP with different encoding methods and activation functions. 
    Our proposed step function encoding achieves the best results, recovering the discrete, sporadic changes in the signal without over-fitting to noise.}
    \label{fig:toy_example}
    \vspace{-3mm}
\end{figure}

To address this problem, we present a novel encoding method based on a step function.
The step function has the desirable property that the output mostly stays constant with respect to the input, except when it changes abruptly at a transition point.
Therefore, we consider using the step function defined below as the encoding function for time $t$:
\begin{equation}
    \stepfunc(t) = \begin{cases}
        0  &  \text{if \;} t \leq u \\
        1  &  \text{if \;} t > u
    \end{cases},
\end{equation}
where $u$ is a learnable parameter representing the 
transition point.
However, $\stepfunc(t)$ is discontinuous and the gradient for $u$ is not well-defined. We therefore use a smooth approximation to $\stepfunc(t)$ to make it differentiable, denoted as $\bar{\stepfunc}(t)$:
\begin{equation}
    \bar{\stepfunc} (t) = \begin{cases}
        \frac{1}{2} \text{exp}(\frac{t-u}{\beta})  & \text{if \;} t \leq u \\
        1 - \frac{1}{2} \text{exp}(\frac{-(t-u))}{\beta})  & \text{if \;}  t > u
    \end{cases},
\label{equ:step_func}
\end{equation}
where $\beta$ is a learnable parameter representing the steepness of the step function. 
In practice, $u$ is randomly initialized from zero to one and $\beta$ is initialized to 0.3. Our encoding method uses a vector of step functions, denoted as $\mulstepfunc(t)$, each with its own learned transition point, 
to express multiple transition points. $u$ and $\beta$ are jointly learned during training.
We experimentally show that we can simply set the dimension of this vector to a large number that exceeds the expected number of scene transitions. 

To illustrate the effectiveness of our proposed encoding function, we compare it with different encoding functions on a toy 1-D fitting experiment in \cref{fig:toy_example}. 
Baseline methods either \emph{overfit} the noise (positional encoding~\cite{mildenhall2020nerf}, Gaussian~\cite{ramasinghe2022beyond}) or \emph{underfit} the discrete, sporadic changes (without encoding, SIREN~\cite{sitzmann2020implicit}).
In contrast, our step function encoding correctly recovers the sharp changes in the signal by approximating real step functions with small 
$\beta$ parameters.
Note that Gaussian and SIREN are used as the activation layer of a network, while our method and positional encoding are used to modulate the input.

\subsection{Implementation Details}
\noindent \textbf{Learning scene appearance with parametric encoding.}
Using an implicit representation to reconstruct a large scene featuring content changes over time requires a large model capacity. 
While we could simply 
increase the size of the MLP, this strategy incurs a linear increase in training and rendering time.
Motivated by Neural Sparse Voxel Fields (NSVF)~\cite{liu2020nsvf}, we adopt a \emph{parametric encoding} which adds additional trainable parameters
for scene appearance to effectively increase model capacity without introducing as much overhead.
Observing that our target man-made scenes often satisfy the Manhattan world assumption, we use a tri-plane structure to arrange additional trainable parameters~\cite{chan2021eg3d}, which we found to be compact and expressive 
in our experiments.
Specifically, we define three learnable feature planes: $\mathbf{E}_{xy}$, $\mathbf{E}_{yz}$, $\mathbf{E}_{xz}$. Each feature plane has a resolution of $D\times D\times B$, where $D$ and $B$ denote the spatial resolution and number of feature channels, respectively. 
 Given a 3D point $\mathbf{p}$, we project it onto three axis-aligned orthogonal planes to obtain $\mathbf{p}_{xy}, \mathbf{p}_{yz}, \mathbf{p}_{xz}$. We can fetch the feature $\mathbf{f}_{xy} = \textbf{interp}(\mathbf{E}_{xy}, \mathbf{p}_{xy})$, where $\textbf{interp}$ is a linear interpolation operation. The same method can be applied to obtain $\mathbf{f}_{yz}$ and $\mathbf{f}_{xz}$. The appearance parametric encoding for $\mathbf{p}$ is defined as the concatenation of $\mathbf{f}_{xy}$, $\mathbf{f}_{yz}$ and $\mathbf{f}_{xz}$.


\medskip
\noindent \textbf{Handling transient objects.}
Learning a scene representation using Internet photos with transient objects may introduce 3D inconsistencies.
To solve this problem, we employ a pretrained semantic segmentation model~\cite{chen2018encoder} to identify pixels of transient objects (e.g., pedestrians) and exclude these pixels during training.
However, a segmentation model may not effectively filter out all transient pixels.
To handle the remaining transient pixels, we use an MLP to predict whether each pixel in each image is a transient object. We learn it using the uncertainty loss, reducing the effect of transient pixels during training models as demonstrated in \cite{chen2022hallucinated}.

\medskip
\noindent \textbf{Other details.}
Our model includes an 8-layer MLP with 256 neurons for each layer as its backbone, and a 4-layer MLP as its appearance head. 
Our model is trained with an initial learning rate of $5e$-$4$, which is reduced to $5e$-$5$ after 800k iterations. 
At each iteration, we randomly sample $1024$ rays from an image.
Following NeRF~\cite{mildenhall2020nerf}, we train our two models using the coarse-to-fine sampling strategy with $64$ and $128$ points for each ray in the coarse and fine levels, respectively.
The model tends to converge after about 800k iterations, which takes about $40$ hours on an RTX 3090 GPU.

\begin{table}
\setlength\tabcolsep{2pt}
\begin{center}
\resizebox{\columnwidth}{!}{
\begin{tabular}{l|cccc}
  \Xhline{3\arrayrulewidth}
    & \textit{Times Square}  & \textit{Akihabara} & \textit{5Pointz} & \textit{The Met}  \\
  \hline
  \# Retrieved images  & 289,794  & 105,445  & 23,628 & 186,663 \\
  \# Calibrated images  & 29,629  & 13,671  & 6,503 & 2,184 \\
  \# Selected images  & 5,965  &    1,078  &  3,521 & 2,127 \\
  \Xhline{3\arrayrulewidth}
  \end{tabular}
}
\caption{\textbf{Dataset statistics.} We collect four scenes for evaluation. For each scene, we first retrieve photos from the Internet, then run COLMAP to calibrate them and reconstruct a sparse point cloud. After calibration, we choose a region of interest from the point cloud and select the corresponding images as input to our method.}
\label{tab:dataset}
\vspace{-7mm}

\end{center}
\end{table}

\section{Experiments}

\subsection{Experimental Setup}
\noindent \textbf{Datasets.}
We collect four scenes from \href{https://www.flickr.com}{Flickr} that include two commercial tourist areas (\textit{Times Square} and \textit{Akihabara}), a graffiti mecca (\textit{5Pointz}), and a museum (\textit{the Metropolitan Museum of Art} aka \emph{the Met}).
The two commercial areas 
feature an array of billboards and other elements that change over time. \textit{5Pointz} is an outdoor space where artists paint graffiti art over time, each piece replacing (or augmenting) a previous one. \textit{The Met} has a varying array of banners and signs advertising different exhibitions. 
For each scene, we run COLMAP to recover camera parameters and a sparse point cloud. Due to the large number of input images, running COLMAP on some of these scenes can take weeks on a cluster with multiple servers. We will release our processed data and data processing scripts as a resource for the community.
Note that whole calibrated scenes, such as \textit{Times Square}, can be excessively large for reconstruction using implicit representations.
Instead, we perform view synthesis experiments on a region of the scene. 
\cref{tab:dataset} summarizes these datasets. 
We include visualizations of the reconstructed models and the selected regions, along with data processing details, in the supplemental material. 

\medskip
\noindent \textbf{Metrics.}
To measure view synthesis quality, we randomly select a few dozen images per scene as a test set. 
To ensure the validity of our evaluation, we visually inspect the images in the test set and remove those that have excessive noise or that do not align with the intended task, such as black and white photos or images that are primarily portraits.
For the remaining test images, there may still be some transient objects present. We manually annotate masks for these objects and ignore them during evaluation.
Following NeRF-W~\cite{martinbrualla2020nerfw}, we use half of the valid pixels of each image to finetune the illumination embedding and the other half for testing PNSR/SSIM/LPIPS metrics.
Please refer to the supplemental material for additional details.
\begin{table}
\setlength\tabcolsep{2pt}
    \begin{center}
    \resizebox{\columnwidth}{!}{
    \begin{tabular}{l|c|cc|ccc}
    \Xhline{3\arrayrulewidth}
      & \multirow{2}{*}{\makecell[c]{ \textfirst{Act.\ func.} / \\ \secondtext{Freq.} \ /  \thirdtext{Dim.}  }} & \multicolumn{2}{c|}{Temporal stability} & \multicolumn{3}{c}{View synthesis quality}  \\ 
      & & Mean $\downarrow$ & Entropy $\downarrow$ & PSNR $\uparrow$ & SSIM $\uparrow$ & LPIPS $\downarrow$ \\
    \hline
   w/o. Time & - & N/A & N/A & 18.18 & 0.572 & 0.461 \\ \hline 
   w/o. Encoding & - & 0.116  & 5.314 &  20.54 &  0.719 & 0.296 \\ \hline
    Learned Latent & - & 11.04 & 5.590 &  20.95  & 0.731  & 0.291 \\ \hline
    \multirow{2}{*}{Activation} & \cellfirst SIREN\cite{sitzmann2020implicit} & 0.101 & 5.552 & 20.54 & 0.696 &  0.330 \\ 
    & \cellfirst Gaussian\cite{ramasinghe2022beyond} & \textbf{0.088} & 5.512 & 20.28 & 0.704 &  0.315 \\ \hline 
    \multirow{3}{*}{\makecell[c]{Positional \\ Encoding}}
     & \cellsecond 5 & 0.247  & 4.936  & 20.71 & 0.731 & 0.288  \\
     & \cellsecond 10 & 5.657  & 5.795   & 20.57 & 0.724 & 0.294  \\
    & \cellsecond 15 &  9.995 & 5.777  & 20.64 &  0.721 & 0.301 \\
    \hline
    \multirow{4}{*}{\makecell[c]{Step Func. \\ Encoding}} & \cellthird 8 & 0.102 & \textbf{1.602} & 20.52 & 0.728 & 0.289 \\
    & \cellthird 16 
     & 0.147 & 2.213 & \textbf{21.32} & \textbf{0.745} &  \textbf{0.274} \\
     & \cellthird  24 & 0.190 & 2.625 & 21.09 & 0.734 & 0.282 \\
     & \cellthird 32 & 0.217 & 2.806 & 21.10 & 0.738 & 0.281 \\
    \Xhline{3\arrayrulewidth}
    \end{tabular}
    }
    \caption{\textbf{Quantitative ablation of the proposed time encoding on \textit{5Pointz}.} 
    \emph{Act.\ func.}~represents the activation function while \emph{Freq.}\ is the frequency of standard positional encoding~\cite{mildenhall2020nerf}. \emph{Dim.}\ is the dimension of the vector size for the learned step functions. 
    } 
    \label{tab:01_ablation}
    \vspace{-7mm}
    \end{center}
\end{table}

\begin{figure*}
    \centering
    \includegraphics[width=\linewidth]{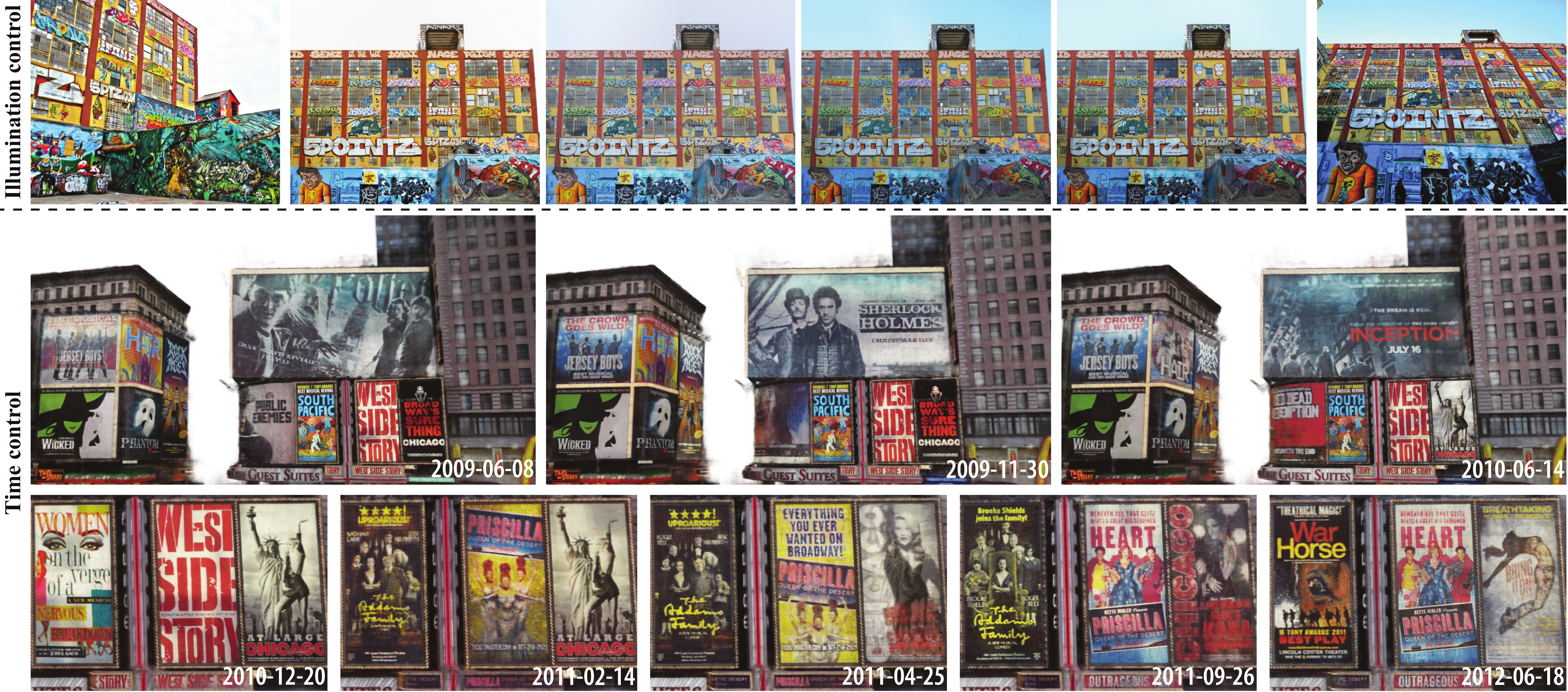}
    \caption{\textbf{Independent control of illumination effects and time.} The top row shows the results of interpolating the illumination embeddings of two real images of \emph{5Pointz}, where the leftmost and rightmost images are real images, and all other images are rendered using our method.  The bottom rows show the results of rendering scenes across time with fixed illumination embeddings (for \emph{Times Square}).}
    \label{fig:control}
    \vspace{-3mm}
\end{figure*}
\subsection{Ablations and Analysis}
\noindent \textbf{Qualitative ablation of the step function encoding.}
An advantage of our method is that it can model abrupt scene-level content changes without overfitting per-image illumination noise.
We compare our method with two baselines: (1) ordinary time input without any encoding, and (2) positional encoding of time~\cite{mildenhall2020nerf} (with a frequency of 15).
We seek to visualize the temporal \emph{stability} of each method.
To do so, given a 4D reconstruction, we first render a video of the scene through time from a fixed viewpoint and with a fixed illumination code (i.e., only content changes). We then compute the mean squared error (MSE) between every two consecutive frames in this video. 
Plots of MSE over time for each method are shown in \cref{fig:main_ablation}. An ideal plot should have large MSE values at sparse points due to abrupt content changes, and zero MSE elsewhere. Our method yields results that exhibit this desired behavior. In contrast, the baseline with no temporal encoding (raw time input) produces smooth transitions across content changes, while positional encoding of time leads to flickering videos.
To further illustrate these behaviors, 
we visualize images around a scene appearance transition point 
in the first row of \cref{fig:main_ablation}.

\medskip
\noindent \textbf{Quantitative ablations and sensitivity analysis.}
\label{par:sensitivity}
In addition to view synthesis quality, we also evaluate the temporal stability of synthesized views under a fixed illumination embedding, which indicates the degree of 
disentanglement between illumination effects and content changes. 
Similar to the ablations discussed above, we measure temporal stability using the statistics of MSE image differences between each two adjacent frames over time, for videos rendered at a fixed viewpoint with fixed illumination embedding.
Specifically, we choose the mean and entropy of these MSE values as our stability metrics. 
Higher mean values indicate significant changes between adjacent frames, corresponding to temporal flicker.
High entropy is 
associated with 
high uncertainty in the distribution.
Considering 
image differences over time, high entropy values indicate that scene content changes are distributed 
throughout time, and are not ``peaky''.
Smaller mean and entropy values indicate that the changes are more concentrated, indicating better modeling of scene content with discrete and sporadic changes.
Please refer to the supplement for additional details on how these metrics are calculated.

We quantitatively ablate our encoding method in terms of 
view synthesis stability and quality in Tab.~\ref{tab:01_ablation}.
We show several baselines and variants: (1) \emph{w/o Time} which does not take time as input, 
(2) \emph{w/o Encoding}, which directly takes raw, unencoded time as input, 
and (3) \emph{Learned Latent}, mapping time to a set of learned latent codes.
We also change the activation function from ReLU to SIREN~\cite{sitzmann2020implicit} and to Gaussian~\cite{ramasinghe2022beyond}, 
which have been shown to have more powerful modeling abilities. 
The baselines of positional encoding with different frequencies are also included.
While many variants achieve reasonable reconstruction quality, 
our method can also achieve both low mean and entropy.

We also provide a sensitivity analysis on the dimension of the vector size of the learned step functions (Dim.) in \cref{tab:01_ablation}. 
The results show that our method can achieve high view synthesis quality once the dimension is $\ge 16$.
Larger dimensions lead to slightly lower temporal coherence, but not a large degradation.
This suggests that we can simply set the number of step functions to a number larger than the number of expected changes in that scene. 
We set Dim.\ to 16 in our experiments for all scenes except Times Square, where we set Dim.\ to 32.

Further qualitative and quantitative ablations of the step function encoding can be found in the supp. material.

\medskip
\noindent \textbf{Application.}
We demonstrate the ability of our method to render plausible and photo-realistic images with controlled time and illumination effects in \cref{fig:control}.

\begin{figure*}
    \centering
    \includegraphics[width=\linewidth]{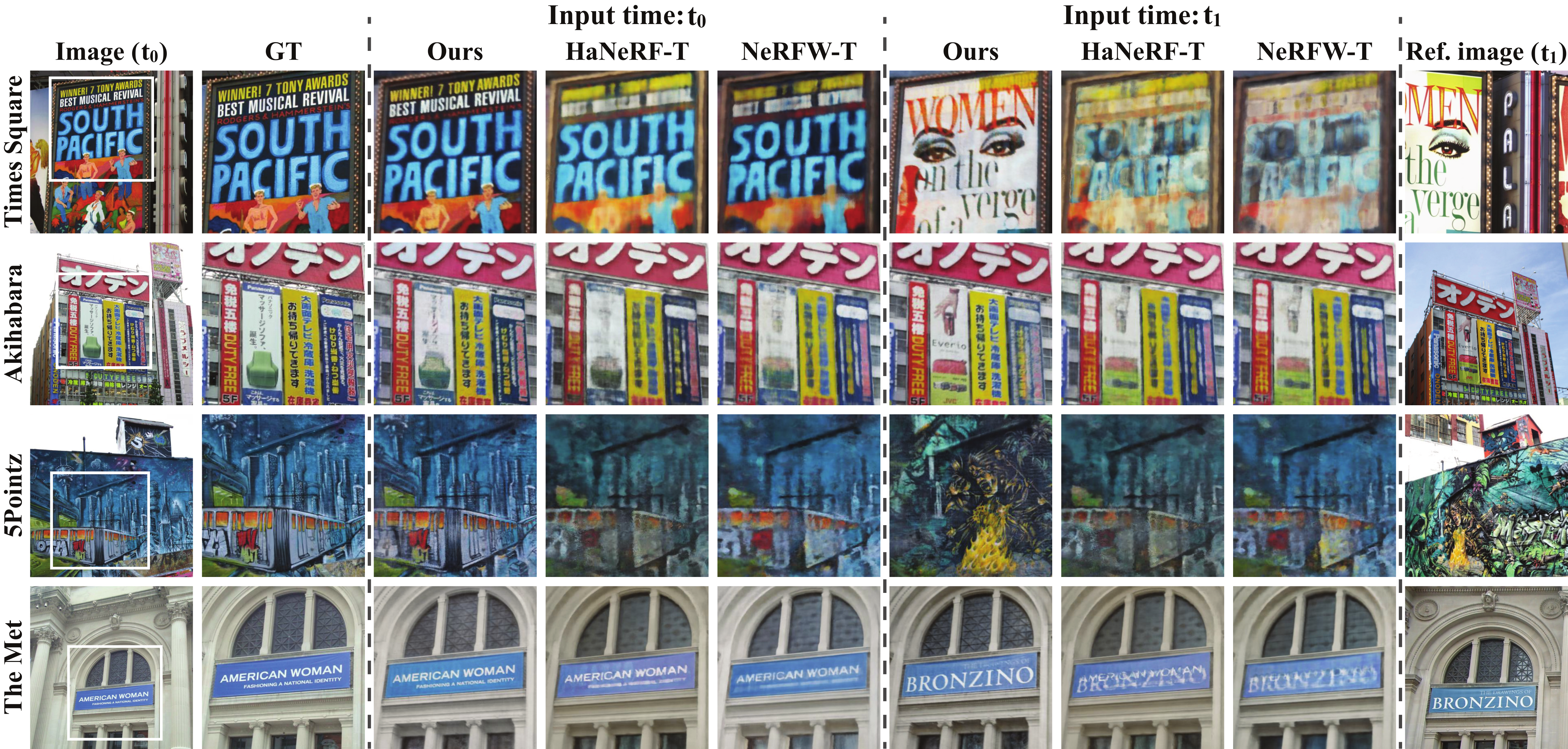}
    \vspace{-4mm}
    \caption{\textbf{Qualitative comparison with the state of the art.} The three column images under \emph{Input time: $\text{t}_{x}$}
    are rendered using timestamps $\text{t}_{x}$  and the viewpoints of the left image.
    Our method renders high-quality images and produces plausible images when changing the input time.
    }
    \label{fig:comp}
    \vspace{-1mm}
\end{figure*}
\begin{table*}
\setlength\tabcolsep{2pt}
    \begin{center}
    \resizebox{\linewidth}{!}{
    \begin{tabular}{l|c|cccc|cccc|cccc|cccc}
    \Xhline{3\arrayrulewidth}
      & \multirow{2}{*}{\makecell{4D view \\ synthesis}} & \multicolumn{4}{c|}{\textit{Times Square}}  & \multicolumn{4}{c|}{\textit{Akihabara}} & \multicolumn{4}{c|}{\textit{5Pointz}} & \multicolumn{4}{c}{\textit{The Met}}  \\ 
      & & Entropy $\downarrow$ & PSNR $\uparrow$ & SSIM $\uparrow$ & LPIPS $\downarrow$ & Entropy $\downarrow$  & PSNR $\uparrow$ & SSIM $\uparrow$ & LPIPS  $\downarrow$ & Entropy $\downarrow$ & PSNR $\uparrow$ & SSIM $\uparrow$ & LPIPS $\downarrow$ & Entropy $\downarrow$  & PSNR $\uparrow$ & SSIM $\uparrow$ & LPIPS $\downarrow$  \\
    \hline
  NeRFW~\cite{martinbrualla2020nerfw} & \XSolidBrush &  N/A & 16.59 &  0.820 & 0.211 & N/A & 17.41  & 0.853  & 0.164 & N/A &17.52&0.545&0.500 & N/A & 23.21 & 0.881 & 0.159 \\
  HaNeRF~\cite{chen2022hallucinated} & \XSolidBrush &  N/A & 15.39 & 0.807   & 0.218 & N/A & 17.50 & 0.860 & 0.160 & N/A &16.82&0.539&0.508& N/A &  22.32 & 0.882 & 0.158  \\
  NeRFW-T&  \Checkmark &  4.990 & 18.71 & 0.847 & 0.190 & 5.483 & 19.16 & 0.874 & 0.140 & 5.768 &19.41&0.611&0.418 & 4.923 & 23.83 & 0.875 & 0.164  \\
  HaNeRF-T & \Checkmark & 4.929 & 17.36 & 0.844 & 0.189 & 5.565 & 18.39 & 0.873  & 0.140  &  5.881  &17.90&0.585&0.445 & 4.943 &22.56 & 0.881 & 0.156 \\
  Ours & \Checkmark & \textbf{3.122} & \textbf{20.87} &\textbf{ 0.894 }& \textbf{0.132}  & \textbf{2.482} & \textbf{20.31} &\textbf{0.902} & \textbf{0.101} &  \textbf{2.213}  &\textbf{21.32}&\textbf{0.745} & \textbf{0.274} & \textbf{2.399} & \textbf{24.07} & \textbf{0.895} & \textbf{0.129}  \\
    \Xhline{3\arrayrulewidth}
    \end{tabular}
    }
    \caption{\textbf{Quantitative comparison with the state of the art.} We augment NeRF-W and HaNeRF to take time as input (*-T). 
    Our method outperforms prior methods across all metrics, demonstrating that our method can better handle such time-varying Internet collections.
    } 
    \label{tab:comp}
    \end{center}
    \vspace{-4mm}
\end{table*}

\subsection{Comparisons with the State of the Art}

We compare to the state-of-the-art methods NeRF-W~\cite{martinbrualla2020nerfw} and HaNeRF~\cite{chen2022hallucinated}, which both reconstruct high-fidelity scene models via implicit representations~\cite{mildenhall2020nerf}. NeRF-W models illumination using per-image embeddings, while HaNeRF models illumination using a CNN module.
However, these methods are designed for static landmarks and cannot handle our test scenes with substantial content changes. We therefore extend these methods by adding time as a network input for fairer comparisons.

We present quantitative and qualitative comparisons with these methods in \cref{tab:comp} and \cref{fig:comp}.
Our method produces lower entropy across all the scenes. In addition to better 
temporal stability, our method also has better view synthesis quality. 
We attribute this to the use of a well-designed appearance parametric encoding. We include ablations of the parametric appearance  encoding in the supplemental material.
To compare the ability of view synthesis through time, we synthesize photos of the same viewpoint at another time as shown in \cref{fig:comp}. 
The step function encoding helps avoid blending artifacts. In contrast, the other methods often exhibit such artifacts when content changes occur, as is evident in the supplemental video.

\section{Conclusion}

We explored the problem of \textit{chronology reconstruction}, aiming to reconstruct and render temporally complex scenes with controlled viewpoint, time, and illumination effects from Internet photos.
We proposed a new neural scene representation equipped with a novel step function encoding to address several challenges, including the entanglement of illumination variation and scene content changes, as well as abrupt scene content changes.
We also collected a new dataset to benchmark this problem. Experiments show that our method exhibits state-of-the-art performance and is capable of producing plausible, stable view synthesis results across time. 
Detailed ablations and analysis were conducted to validate our proposed components.

\medskip
\noindent \textbf{Limitations and future work.} 
Our method takes as input Internet photos with timestamps. Inaccurate timestamps may hinder the training process, and Internet photos that do not have timestamps cannot be utilized for training. 
Exploring how to simultaneously predict timestamps is an interesting avenue for future work. 
In addition, some urban scenes such as Times Square have billboards that display videos (not still images), which are difficult for our method to reconstruct, as their content has high temporal frequency is not well supported by other images in the collection.

\medskip
\noindent \textbf{Acknowledgements.}
  The authors would like to acknowledge support from Information Technology Center and State Key Lab of CAD\&CG, Zhejiang University. 
\newpage
\newpage
{\small
\bibliographystyle{ieee_fullname}
\bibliography{egbib}
}

\newpage
\appendix
In the supplementary material, we present more implementation details, results, and discussion.

\section{Method Details}

\subsection{Data Processing Details}
Given Internet photos of a landmark, one way to obtain camera calibration and scene structure is to run COLMAP with built-in acceleration techniques such as Vocabulary Tree Matching and Hierarchical Mapper. However, given the huge number of Internet photos (up to 300K) in our problem, this process requires very high computing resources and up to several TB of memory.
To efficiently calibrate 300K images, we design a scheme that can utilize multiple GPU machines in parallel, while solves the problem of computing resources and memory requirements.
Specifically, we first randomly divide the pictures into several parts that partially overlap, and then run COLMAP for each part on a GPU separately and in parallel. We then merge these models using the overlapping registered images.

The whole calibrated scene may be too big to reconstruct for the implicit representation. Instead of reconstructing the whole scene, we attempt to reconstruct a piece of the scene. Specifically, given a SFM sparse point cloud of the scene, we can select a interested region from it using MeshLab. Then we can obtain the images, which are registered to the model using the points in the selected interesting region.

\begin{figure*}
    \centering
    \includegraphics[width=\linewidth]{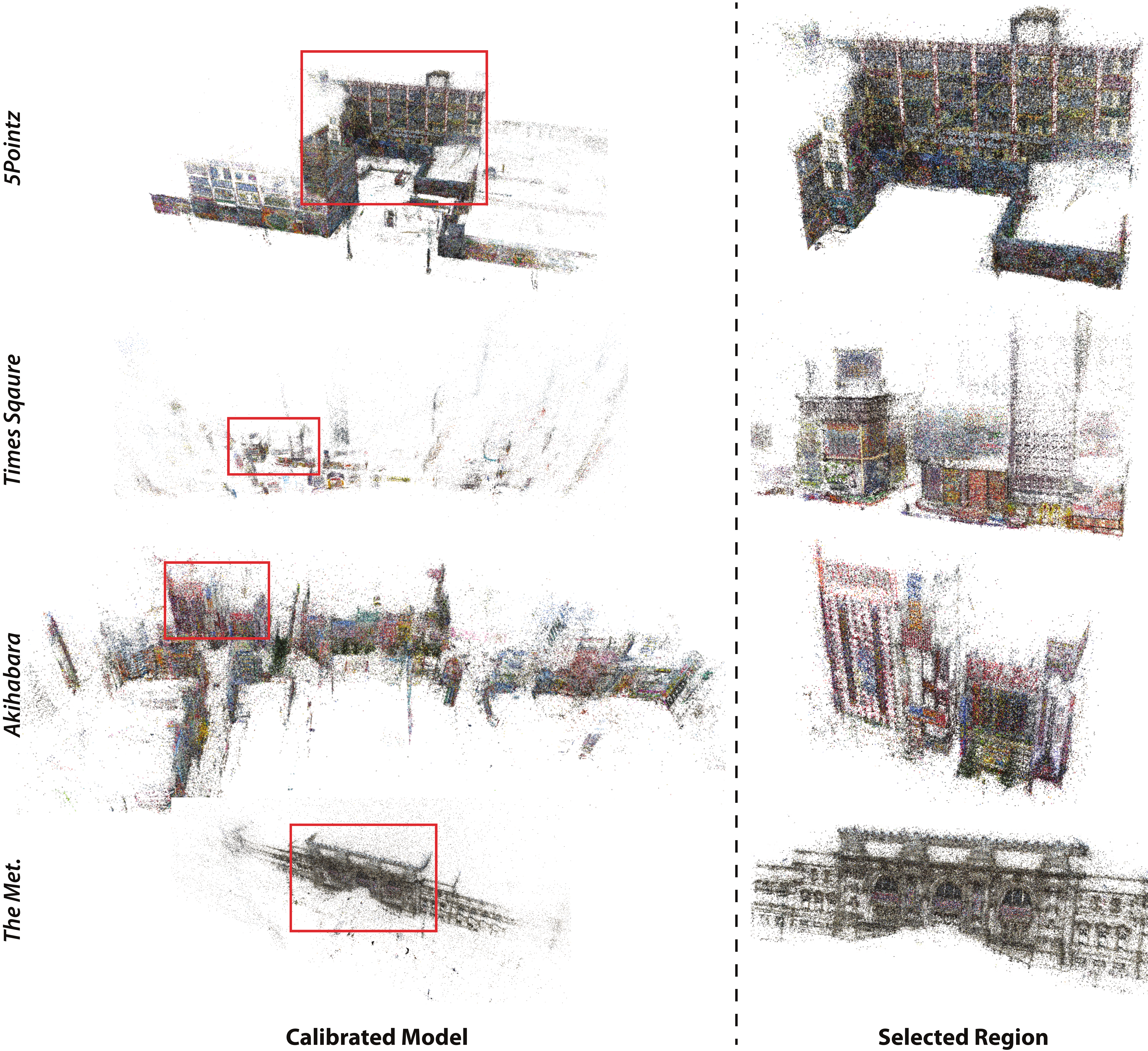}
    \caption{\textbf{Scene scale visualization.} The left column images are the sparse point cloud of the COLMAP model. We select an interesting region from the sparse point cloud using MeshLab.
    }
    \label{fig:scenescale}
\end{figure*}

Please see \cref{fig:scenescale} for SfM results and our selected regions.

\subsection{Sky Modeling}
Given an outdoor scene, we can use the SFM sparse point cloud to define the near and far planes for sampling points to optimize our representation.
For the sky part, this kind of sampling tends to recover a cloud above the building, causing artifacts when we conduct view synthesis over a large angle range due to wrong geometry.
To address the issue, we use a spherical radiance map proposed in \cite{hao2021gancraft} to model the sky.
Specifically, we map the view direction and the image illumination embedding to the sky color using an MLP and then use alpha blending to obtain the final image color.

\subsection{Model Architecture Details}
\begin{figure*}
    \centering
    \includegraphics[width=\linewidth]{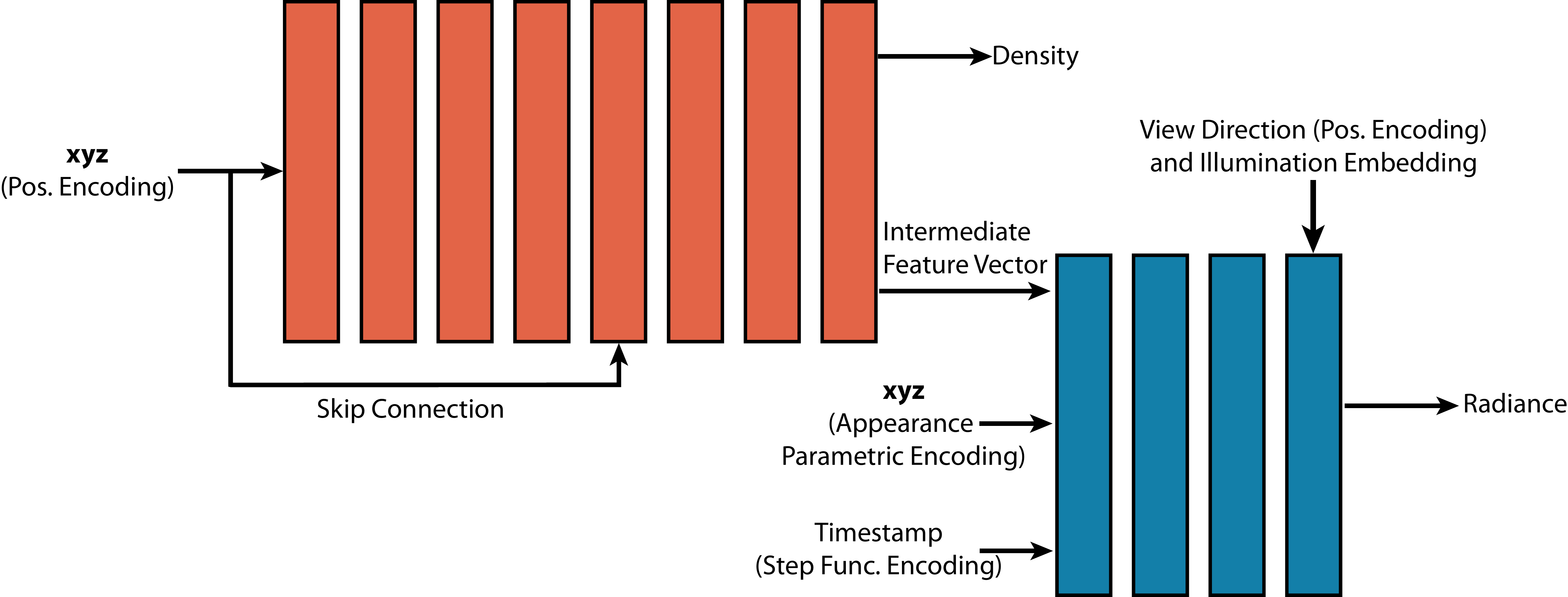}
    \caption{\textbf{Model Architecture.} \textit{Step Func. Encoding} and \textit{Pos. Encoding} represent \textit{Step Function Encoding} and \textit{Positional Encoding}, respectively. Each layer of the MLP has 256 neurons. We use ReLU as our activation function. }
    \label{fig:architecture}
\end{figure*}

\cref{fig:architecture} shows our model architecture. 
Given a 3D point coordinate $\mathbf{xyz}$ and timestamp, along with an 
illumination embedding and view direction, we use an MLP to predict
its radiance and density. 
For simplicity, we assume that the scene geometry is mostly constant, and only the appearance changes over time. Therefore, we use an MLP (colored in orange in \cref{fig:architecture}) to model the geometry and use an MLP (colored in blue in \cref{fig:architecture}) to model time-varying appearance.

\begin{figure}
    \centering
    \includegraphics[width=\linewidth]{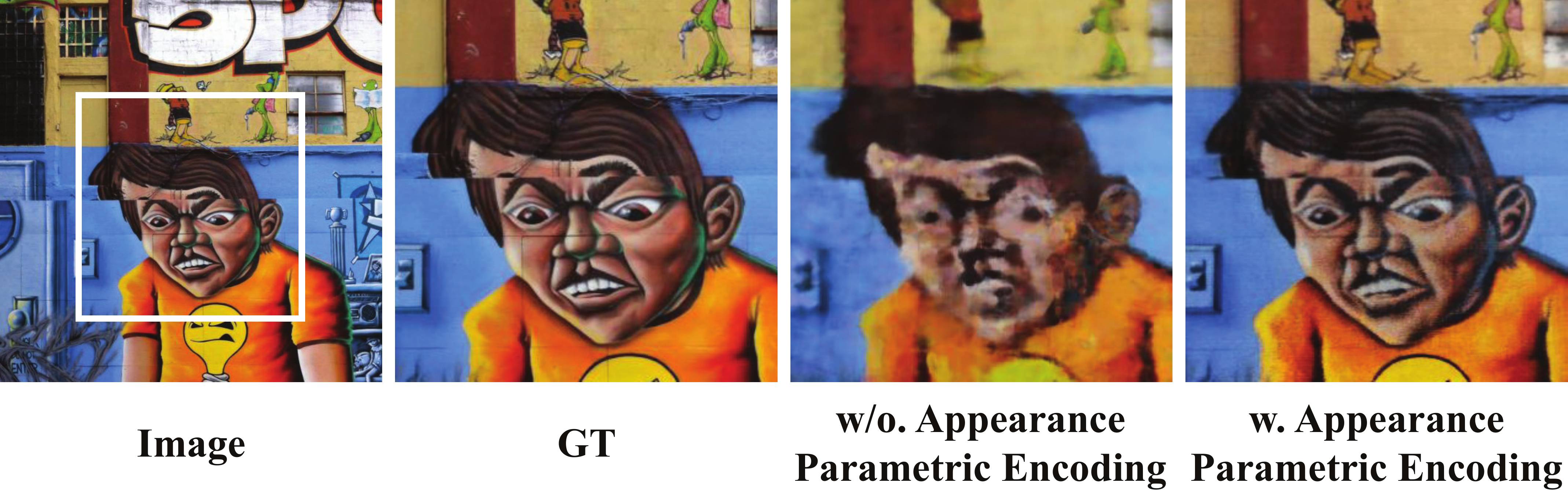}
    \caption{\textbf{Ablation on the appearance parametric encoding.} As illustrated above, using a parametric encoding significantly affects the rendering quality. }
    \label{fig:parametric_encoding}
\end{figure}

\subsection{Entropy Details.}
Our technical contribution lies in the step function encoding, which enables us to model abrupt \textit{scene-level} content change without overfitting \textit{per-image} illumination variation. We use the view synthesis stability (Entropy) along the time to quantify it. Specifically, we first render a video with a fixed viewpoint and a fixed illumination embedding. Then we compute the image difference (MSE) over two consecutive frames of this video to obtain the MSE over time. 
The distribution of MSE along the time can be regarded as the distribution of content changes over time. 
The ideal distribution should be concentrated into several points.
Given a distribution $P$, which takes values in the alphabet $\mathcal{P}$ and is distributed by $p: \mathcal{P} \rightarrow [0,1]$ (the normalized MSE over time), its entropy $E$ can be computed as:
\begin{equation}
    E(P) = \sum_{x_i \in \mathcal{P}} -p(x_i)\text{log\;} p(x_i).
\end{equation}
When $x$ is equal to zero or one, the function $y=-xlog(x)$ is zero, while this function is positive when $x$ is between zero and one. 
An ideal concentrated distribution has zero entropy. 

\begin{figure}
    \centering
    \includegraphics[width=\linewidth]{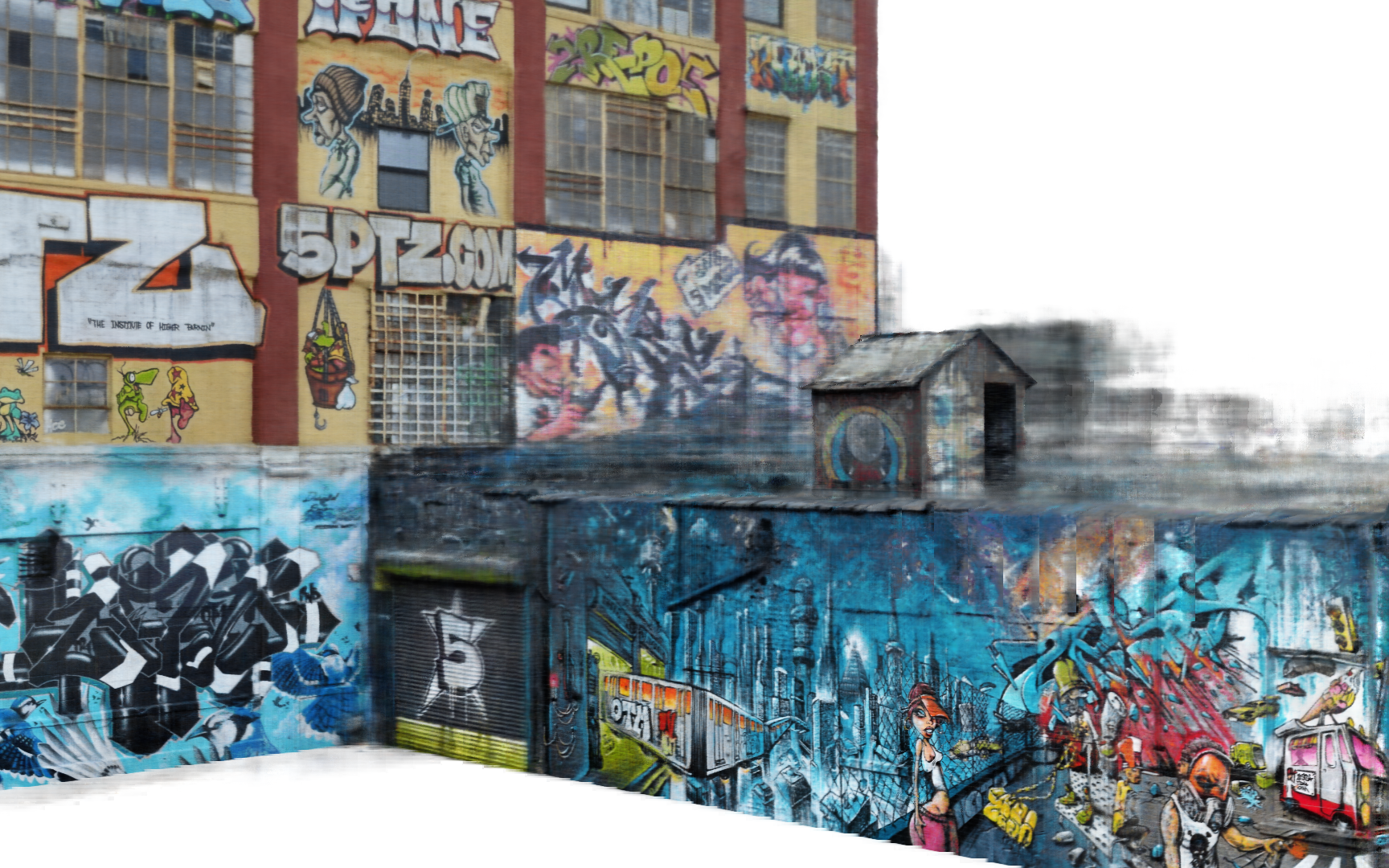}
    \caption{\textbf{Our method cannot reconstruct blind spots well.}  People cannot see the top of this small platform when standing on the ground to take pictures.}
    \label{fig:artifacts}
\end{figure}

\section{Additional Results}

\subsection{Ablation on the Step Func. Encoding}
\begin{figure*}
    \centering
    \includegraphics[width=\linewidth]{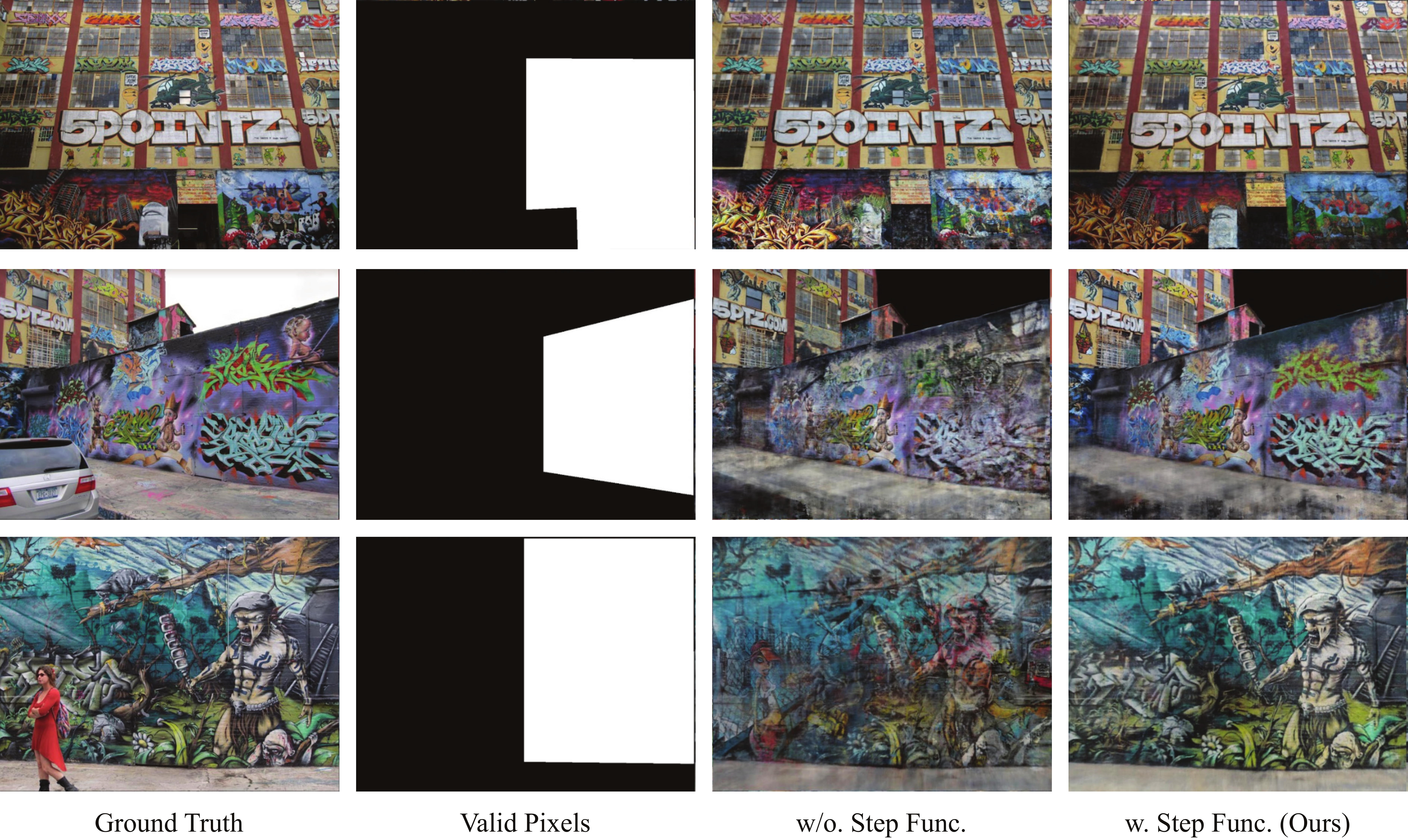}
    \caption{\textbf{Qualitative ablation on the step function encoding.} We focus on the evaluation of manually annotated valid pixels for reasonable metric computation. 
    It can be observed that blending artifacts appear when step function encoding (Step Func.) is not used. This demonstrates the effectiveness of Step Function Encoding. 
    }
    \label{fig:moreablation}
\end{figure*}

To further demonstrate the advantages of step function encoding (in addition to the ablation on the step function encoding we present in the main paper), we carefully selected a train/test split for 5Pointz and evaluated it by visual inspection. This test split contains a large number of images captured at the time of scene content changes. Without using the step function encoding, there would be noticeable blending artifacts in such a test split. Specifically, the results of using step function encoding versus not using it are 19.19/0.715/0.307 and 18.07/0.678/0.345 (PSNR/SSIM/LPIPS), respectively. We present qualitative results on \cref{fig:moreablation}.

\subsection{Ablation on the Parametric Encoding} 
We find that the parametric encoding for appearance modeling significantly affects the rendering quality. 
Specifically, models with and without proposed parameter encoding produce 19.48/0.612/0.423 and 21.32/0.745/0.274 on \textit{5Pointz} in terms of PSNR/SSIM/LPIPS metrics, respectively. We provide a visual ablation on \cref{fig:parametric_encoding}.

\subsection{Ablation on the Static-geometry Assumption}
We run an ablation study on \textit{5Pointz} testing the effect of allowing dynamic geometry, and the results indicate a slight decrease in performance (20.82/0.729 vs.\ 21.32/0.744 in terms of PSNR/SSIM). Note that all baselines we evaluate assume static geometry, to ensure a fair comparison.


\subsection{Scalability of Step Function Encoding}
\begin{figure}
    \centering
    \includegraphics[width=0.89\columnwidth]{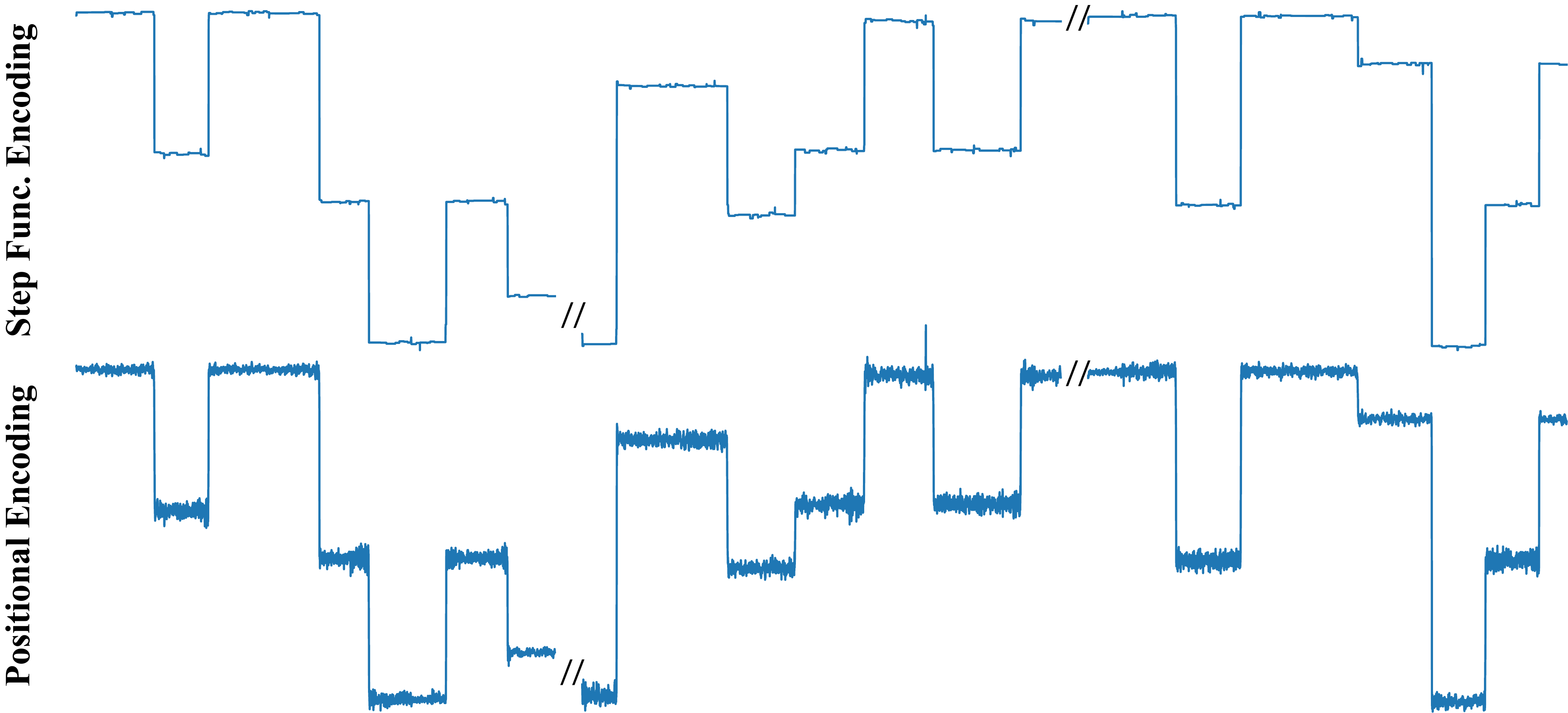}
    \vspace{-1mm}
    \caption{\textbf{An extension of the experiment in Fig.\ 3 of main paper. } The input noisy signal has 96 transitions, and the dimension of the step function encoding is set to 128. This figure shows fragments of selected transitions. Compared to positional encoding, our method exhibits better temporal stability.}
    \vspace{-1mm}
    \label{fig:transitions}
\end{figure}

\cref{fig:transitions} shows that our method works for synthetic signals with nearly 100 transitions, indicating the scalability of our approach. 
Specifically, the synthetic signals have multiple dimensions, with each dimension consisting of a randomly segmented and discretized value. Such signals are similar to real-world scenarios, where the value in each dimension can be analogized to a billboard in a real-world scene.

\section{Discussion}



\subsection{Time Period Selection}
We reconstructed 4 scenes in the period 2009---2013. We chose this period because we found that Flickr was actively used during this period, which provides us with lots of photos. After this period, Flickr became less popular, perhaps due to the emerging social media platforms. 
For example, Flickr has nearly 80,000 pictures about \textit{Akihabara} in 2010-2013, but there are less than 10,000 pictures about \textit{Akihabara} in the last 4 years.
Future datasets should also consider these social platforms, if interested in representing more recent years.

\subsection{Additional discussion on transient object detection}
\textit{Why does transient object detection not interfere with our goal to disentangle per-image changes from scene-level changes?} 
The appearance of a scene at a certain stage is usually observed by multiple images, while transient objects are not.
Uncertainty loss tends to increase the uncertainty of regions where the model cannot learn (e.g., regions of transient objects), while minimizing uncertainty in other regions. 
The model can more easily learn scene appearance changes in a 3D coherent way to fit regions with scene appearance changes, rather than predicting these regions with high uncertainty. 

\subsection{Additional Discussion on Artifacts}
In addition to limitations in Sec.5 of the main paper, our method also has the limitation of not being able to reconstruct regions with very few observations. This problem is relatively common for landmark reconstruction from Internet photos. We include some visualization results in \cref{fig:artifacts}.

\end{document}